\documentclass[conference]{IEEEtran}
\usepackage{cite}
\usepackage{amsmath,amssymb,amsfonts}
\usepackage{algorithmic}
\usepackage{graphicx}
\usepackage{textcomp}
\usepackage{xcolor}
\usepackage{multirow}

\def\BibTeX{{\rm B\kern-.05em{\sc i\kern-.025em b}\kern-.08em
		T\kern-.1667em\lower.7ex\hbox{E}\kern-.125emX}}

\begin{document}
	
	\title{Handwritten Indic Character Recognition using Capsule Networks}		
	\author{
		\IEEEauthorblockN{Bodhisatwa Mandal\IEEEauthorrefmark{1},Suvam Dubey\IEEEauthorrefmark{1}, Swarnendu Ghosh\IEEEauthorrefmark{1}, RiteshSarkhel\IEEEauthorrefmark{2}, Nibaran Das\IEEEauthorrefmark{1}}
		\IEEEauthorblockA{\IEEEauthorrefmark{1}Dept. of CSE, Jadavpur University, Kolkata, 700032, WB, India.
			\\\{bodhisatwam,suvamdubey\}@gmail.com, {swarnendughosh.cse.rs,nibaran.das}@jadavpuruniversity.in}
		\IEEEauthorblockA{\IEEEauthorrefmark{2}Dept. of CSE, Ohio State University, Columbus, OH 43210, USA
			\\ sarkhelritesh@gmail.com}
	}

	\maketitle
	
	\begin{abstract}
	Convolutional neural networks(CNNs) has become one of the primary algorithms for various computer vision tasks. Handwritten character recognition is a typical example of such task that has also attracted attention. CNN architectures such as LeNet and AlexNet have become very prominent over the last two decades however the spatial invariance of the different kernels has been a prominent issue till now. With the introduction of capsule networks, kernels can work together in consensus with one another with the help of dynamic routing, that combines individual opinions of multiple groups of kernels called capsules to employ equivariance among kernels. In the current work, we have implemented capsule network on handwritten Indic digits and character datasets to show its superiority over networks like LeNet. Furthermore, it has also been shown that they can boost the performance of other networks like LeNet and AlexNet.
	\end{abstract}
	

\section{Introduction}
It has been two decades since the first convolutional neural networks was introduced in 1998 ~\cite{lenet} for handwritten digit classification problem. Since then computer vision has matured a lot in terms of both the complexity of the architectures as well as the difficulty of the challenges they address. Many works have been introduced in later years to address challenges like object recognition~\cite{alexnet}. However through all these years one principal issue was yet to be addressed. Convolutional Neural Networks by its nature employ invariance of features against their spatial position. As the kernels that represent specific features are convolved throughout the entire image, the amount activation is position invariant. The activations across different kernels do not communicate with each other and hence their outputs are spatially invariant. We have intuitively developed the skill to analyze relative positions of various parts of an object. To learn these relations the capsule networks were proposed ~\cite{routing}. In short capsule networks consist of a group of kernels that work together and pass information to next layers through a mutual agreement that is achieved by dynamic routing of information during the forward pass. In our experiments, our goal is to analyze the performance of these networks for some Indic digit and character datasets. We have used three of the most popular Indic digits dataset namely,Devanagari digits, Bangla digits, and Telugu digits. While these have only 10 classes, the two character datasets, namely, Bangla basic character and Bangla compound characters have 50 and 199 classes respectively. There have been many works in Indic datasets using CNNs before ~\cite{indic3,indic6}. In our experiments, the performance of the capsule networks are compared with respect to LeNet and AlexNet. To show that the capsule network learns unique concepts, we have combined it with other networks to show a boost in performance. In the next section, a refresher is provided as to the basics of a simple CNN. In section 3 it is shown how the capsule network evolves over the simple CNN along with explanations regarding its internal mechanisms. In section 4, the experimentations and results are discussed and finally concluding in section 5.
\begin{figure*}[htbp]
	\centering
	\includegraphics[width = 0.75\textwidth]{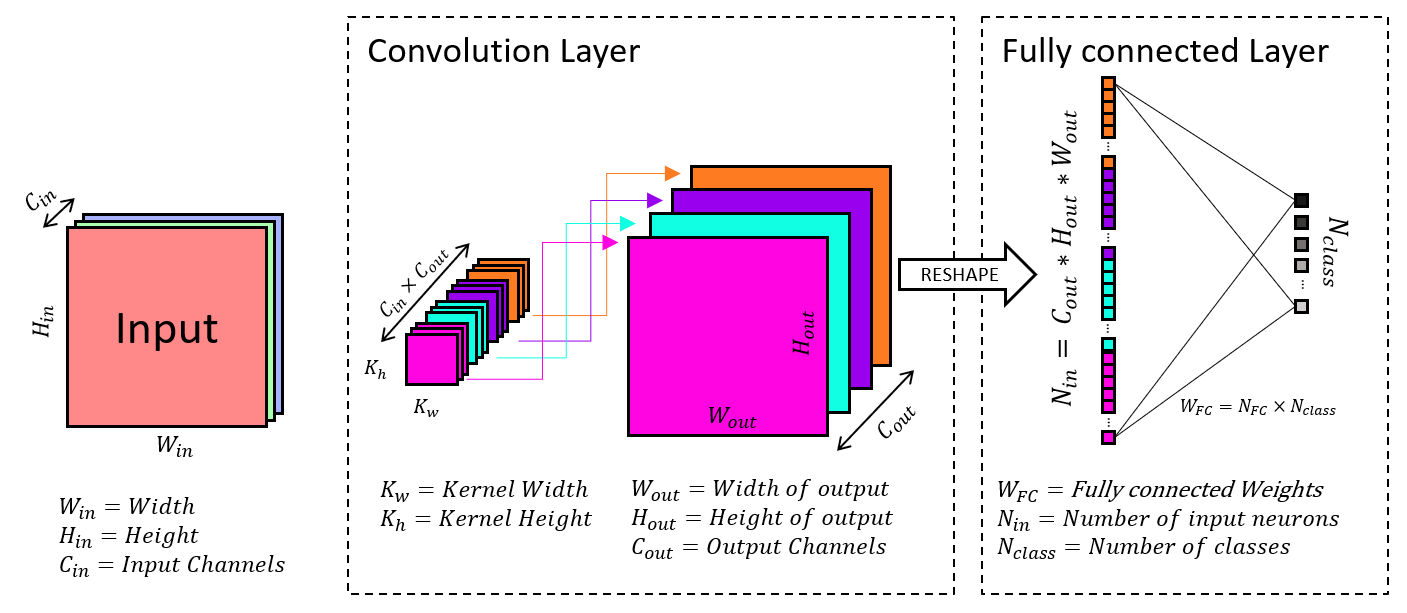}
	\caption{A schematic diagram of a sample convolutional network(Pooling Layers omitted to keep the architecture analogous with respect to fig 2.}
\end{figure*}
\section{CNN Refresher}
Convolutional neural networks are typically designed as a series of 2-d convolution and pooling operations along with non-linear activations in the middle followed by a fully connected network for classification. A 2-d convolution is performed by convolving a kernel $K$ over an input of size $C_{in} \times H_{in} \times W_{in}$ where $H_{in},W_{in}$ and $C_{in}$ are height, width and number of input channels of the input . A kernel convolution on such an input should be of a shape $C_{in} \times K_h \times K_w$. Here $C_{in}, K_h, K_w$ are the depth, height and width of the kernel. Note that the depth of the kernel is equal to the input number of channels. If we use $C_{out}$ number of such kernels, then the output tensor generated shall be of the shape $C_{out}\times H_{out}\times W_{out}$. The height $H_{out}$ and width $W_{out}$ are dependent on factors like input height $H_{in}$, input width $W_{in}$, stride of the kernel and the padding of the input. Convolutions are typically followed by non-linear activations such as a sigmoid, Tanh, or a rectified linear units. Pooling operations normally take a small region of the input and compresses it to a single value by taking either maximum(max pooling) or average(average pooling) of the corresponding activations. This reduces the size of the activation maps. Hence when kernels convolve over this tensor, it actually corresponds to a larger area in the original image. After a series of convolutions, activations and pooling, we obtain a tensor signifying the extracted features of the image. This tensor is flattened to form a linear vector of shape $C''*H''*W'' \times 1$, which can be fed as an input to a fully connected network. Here $C'',H''$ and $W''$ are the depth, height, and width of the tensor to be flattened. The total number of neurons in this layer $N_{FC}$ is $C''*H''*W''$. At the end of the fully connected network we get a vector of size $N_{class}\times1$ that corresponds to the output layer. A loss such as mean-square error, or cross entropy, or negative log likelihood is computed which is then back-propagated to update the weights using optimizers like stochastic gradient descent or adaptive moments. A schematic diagram is shown in fig.1, with a typical convolution operation followed by a fully connected layer to perform classification. Layers such as pooling and non-linearities are not shown to keep simplicity and to keep the diagram analogous to fig. 2.

\section{The Capsule Network}
The primary concern with CNNs are that the different kernels work independently. If two kernels are trained to activate for two specific parts of an object they will generate the same amount activations irrespective of the relative positions of the object. Capsule networks brings a factor of agreement between kernels in the equation. Subsequent layers receive higher activations when kernels corresponding to different parts of the object agree with the general consensus. The capsule network proposed in ~\cite{routing} consist of two different capsule layers. A primary capsule layer that groups convolutions to work together as a capsule unit. This is followed by a digit capsule layer that is obtained by calculating agreement among different capsules through dynamic routing. A schematic diagram of capsule network is provided in fig. 2. The diagram does not represent the actual architecture proposed in the original work ~\cite{routing}, rather it demonstrates a primary and digit capsule layer. The diagram drawn is kept analogous to a typical CNN shown in fig.1 to highlight the major differences.
\begin{figure*}[htbp]
	\centering
	\includegraphics[width = 0.75\textwidth]{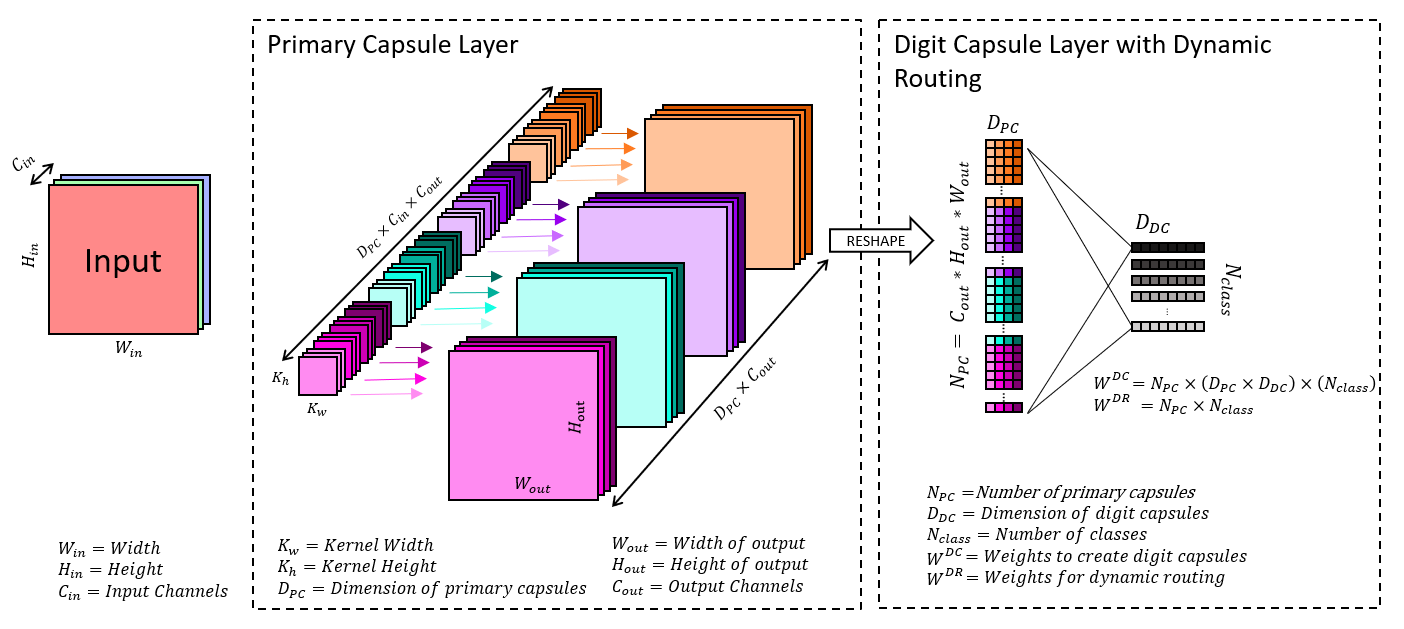}
	\caption{A schematic diagram of capsule network~\cite{routing} demonstrating primary and digit capsule layers.}
\end{figure*}
\subsection{Primary Capsules}
The capsule network starts with typical convolution layer that converts the input image into a block of activations. This tensor is fed as an input into the primary capsule layer. If the number of channels in this input is $C_{in}$ and the desired dimension of primary capsules is $D_{PC}$ then the shape of one kernel is $D_{PC} \times C_{in} \times K_h \times K_w$. $K_h$ and $K_w$ are the height and width of the kernel. With $C_{out}$ number of such kernels we shall get an output of shape $C_{out} \times D_{PC} \times H_{out} \times W_{out}$, where The height $H_{out}$ and width $W_{out}$ is dependent on factors like input height $H_{in}$, input width $W_{in}$, stride of the kernel and the padding of the input. Unlike normal convolutions, where each activation tensor had a depth of $1$, the depth of the activations in primary capsules is $D_{PC}$. The total number of primary capsules $N_{PC}$ is $C_{out}*H'*W'$. Before passing to the next layer this tensor is reshaped into $N_{PC} \times D_{PC}$. 
\subsection{Digit Capsules}
Normally output layers in a fully connected network is of the shape $N_{class}\times 1$, where $N_{class}$ is the number of classes. The capsule network replaced the output layers with a digit capsule layer. Each class is represented by a capsule of dimension $D_{DC}$. Hence we get a digit capsule block of shape $N_{class}\times D_{DC}$. By calculating the L2 Norm of each row we get our output layer of shape $N_{class} \times 1$. The values of digit capsules are calculated by dynamic routing between primary capsules.
\subsection{Dynamic routing}
The dynamic routing~\cite{routing} is computed to obtain the digit capsules from the primary capsules. Two different types of weights are required to perform dynamic routing. Firstly we need the weights to calculate individual opinions of every capsule. These weights, $W^{DC}$ are normally trained using back-propagation. If $i \in [1,N_{PC}]$ is the index of the primary capsules of $D_{PC}$ dimension and $j \in[1,N_{class}]$ is the index of the digit capsules of $D_{DC}$ dimension. $W^{DC}_{ij}$ is of shape $D_{PC}\times D_{DC}$. The individual opinion of $i$ regarding the digit capsule $j$ is given by,
\begin{equation}
\hat{u}_{j|i} = u_i W^{DC}_{ij},
\end{equation}
where $u_i$ is the $i-th$ primary capsule. So for each capsule $i$ we get an individual digit capsule block of shape $N_{class}\times D_{DC}$. The second type of weight can be called the routing weights ($W^{DR}$). The routing weights are used to combine these individual digit capsules to form the final digit capsules. These routing weights are updated on during the forward pass based on how much the individual digit capsules agree with the combined one. The routing weight matrix $W^{DR}$ is of the shape of $N_{PC}\times N_{class}$. During each forward pass the routing weights are first initialized as zeros. The coupling coefficients $c_{ij}$ is given by,
\begin{equation}
c_{ij} = \frac{\exp(W^{DR}_{ij})}{\sum_k \exp(W^{DR}_{ik})}.
\end{equation}
The coupling coefficients are used to combine the individual digit capsules and form the combined digit capsule. The $j-th$ combined digit capsule $s_j$ is given by,
\begin{equation}
s_j = \sum_i {c_{ij} \hat{u}_{j|i}}
\end{equation} 
A squashing function stretches the values of $s_j$ such that bigger values go close to one and lower values go close to zero. The squashed combined digit capsule $v_j$ is given by,
\begin{equation}
v_j = \frac{||s_j||^2}{1+||s_j||^2}\frac{s_j}{||s_j||}.
\end{equation}
The agreement between individual digit capsules $\hat{u}_{j|i}$ and the squashed combined digit capsules $v_j$ can be calculated using a simple dot product. The more the value of the agreement the more preference is awarded to the corresponding capsule $i$ in the next routing iteration. This is obtained by updating the $W^{DR}_{ij}$ as,
\begin{equation}
W^{DR}_{ij} = W^{DR}_{ij} + \hat{u}_{j|i}.v_j.
\end{equation}
Equations 2-4 are repeated for a specific number of routing iterations to perform iterative dynamic routing of opinions of primary capsules to form the digit capsule.
\subsection{Loss Function}
The loss function used for capsule networks is a marginal loss for the existence of a digit. The marginal loss for digit $k$ is given by,
\begin{multline}
L_k = T_k\ max(0,m^{+} - ||v_k||)^2 + \\
	 \lambda\ (1-T_k)\ max(0,||v_k||-m^{-})^2.
\end{multline}
Here, $T_k = 1$  iff a digit of class $k$ is present. The upper and lower bounds $m^{+}$ and $m^{-}$ are set to 0.9 and 0.1 respectively. $\lambda$ is set as 0.5
\subsection{Regularization}
Proper regularization of a network is essential to stop models from over-fitting the data. In case of capsule networks a parallel decoder network is connected with the obtained digit capsules as its input. The decoder tries to reconstruct the input image. A reconstruction loss is also minimized along with margin loss so that the network does not over-fit the training set. However the reconstruction loss is scaled down by a factor of 0.0005 so that the margin loss is not dominated.

\section{Experimentations and Results}
Our experiments focus on the implementation of capsule networks for handwritten Indic digits and character databases. The results have been compared with other famous CNN architectures like LeNet and AlexNet. While LeNet was built for smaller problems like digit classification, AlexNet was intended for much more complicated data like the ImageNet. The input image is resized to the native size supported by the network that is $28\times28$ for capsule networks, $32\times 32$ for LeNet and $227\times 227$ for AlexNet. In total 7 models have been tested on 5 datasets. Firstly, the basic LeNet, AlexNet and capsule network was tested. Second set of experiments involved an ensemble of two of the three networks using a probabilistic averaging. Finally all the three networks were combined by averaging the output probability distribution. All the results are tabulated in table 2.
\begin{figure*}[htbp]
	\label{fig:3}
	\centering	
	\includegraphics[width = 0.75\textwidth]{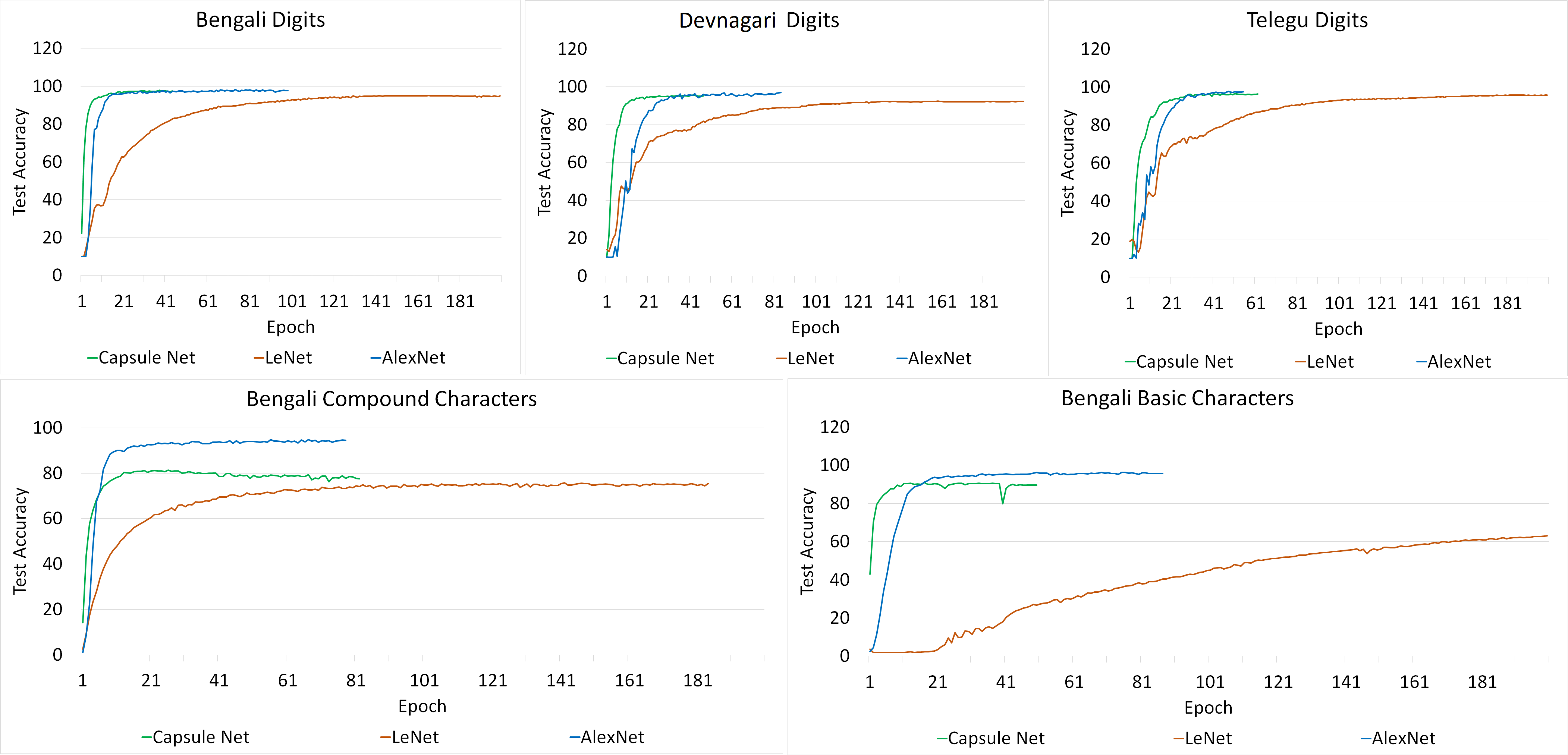}
	\caption{Test accuracy vs number of training epochs for character datasets.}
\end{figure*}

\subsection{Datasets}
We have used five datasets for our experiments. Firstly we have Indic handwritten digit databases(CMATERdb) \footnote{https://code.google.com/archive/p/cmaterdb/} in three scripts that is Bangla(CMATERdb 3.1.1), Devanagari(CMATERdb 3.2.1) and Telugu(CMATERdb 3.4.1). These are a typical 10 class problems to primarily challenge the performance of LeNet. Subsequently, the character databases namely, Bangla basic characters(CMATERdb 3.1.2) and Bangla compound characters(CMATERdb 3.1.3.3) give us a 50 class and a 199 class problem to deal with. The description of the datasets are given below. All the datasets were split into train and test set in the ratio 2:1. The accuracies provided are with respect to the best model in terms of training accuracy.
\begin{table*}[htbp]
	\centering
	\caption{Specifications for the capsule network with respect to a single channel input and ten class output}
	
	\resizebox{0.7\textwidth}{!}{%
		\begin{tabular}{|l|l|c|c|c|c|c|c|c|r|r|}
			\hline
			\textbf{Network} & \multicolumn{1}{c|}{\textbf{\begin{tabular}[c]{@{}c@{}}Type of \\ Layer\end{tabular}}} & \multicolumn{1}{c|}{\textbf{\begin{tabular}[c]{@{}c@{}}Kernel \\ Size\end{tabular}}} & \multicolumn{1}{c|}{\textbf{Stride}} & \multicolumn{1}{c|}{\textbf{Padding}} & \multicolumn{1}{c|}{\textbf{\begin{tabular}[c]{@{}c@{}}Input \\ Channels\end{tabular}}} & \multicolumn{1}{c|}{\textbf{\begin{tabular}[c]{@{}c@{}}Output \\ Channels\end{tabular}}} & \multicolumn{1}{c|}{\textbf{\begin{tabular}[c]{@{}c@{}}Input \\ Capsule \\ Dimension\end{tabular}}} & \multicolumn{1}{c|}{\textbf{\begin{tabular}[c]{@{}c@{}}Output \\ Capsule \\ Dimension\end{tabular}}} & \multicolumn{1}{c|}{\textbf{\begin{tabular}[c]{@{}c@{}}Weights \\ + \\ Biases\end{tabular}}} & \multicolumn{1}{c|}{\textbf{\begin{tabular}[c]{@{}c@{}}Routing \\ Weights\end{tabular}}} \\ \hline \hline
			& Convolution & 9 & 1 & 0 & 1 & 256 & NA & NA & 20,992 & NA \\
			& Primary Caps & 9 & 2 & 0 & 256 & 32 & 1 & 8 & 5,308,672 & 0 \\
			\textbf{Capsule} & Digit Caps & NA & NA & NA & 1,152 & 10 & 8 & 16 & 1,474,560 & 11,520 \\
			\textbf{Networks} & Decoder FC & NA & NA & NA & 160 & 512 & NA & NA & 82,432 & NA \\ 
			& Decoder FC & NA & NA & NA & 512 & 1,024 & NA & NA & 525,312 & NA \\
			& Decoder FC & NA & NA & NA & 1,024 & 784 & NA & NA & 803,600 & NA \\ \cline{2-11} 
			\textbf{} & \multicolumn{8}{l|}{\textbf{Total number of parameters}} & \multicolumn{2}{r|}{\textbf{8,227,088}} \\ \hline
		\end{tabular}%
	}	
\end{table*}
\subsection{Architecture and Hyperparameters}
The capsule network has been used as it has been proposed in ~\cite{routing}. The performance is compared with respect to LeNet and AlexNet. The specifics of the capsule network architecture is provided in table 1. The LeNet was primarily built for MNIST digit classification with only around 61K trainable parameters. The AlexNet has around 57 million trainable parameters so that it can tackle harder problems. Like LeNet, the capsule network was also proposed for MNIST digit classification, however it is much more robust. It has around 8.2 million parameters out of which around 11K parameters are trained on the runtime by dynamic routing. All the provided statistics is with respect to a single channel input of native input size and a 10 class output. All networks are optimized with Adam optimizer with an initial learning rate of 0.001, eps of 1e-08 and beta values as 0.9 and 0.999. The experiments were carried out using a Nvidia Quadro P5000 with 2560 CUDA cores and 16 GB of VRAM.

\begin{table}[h]
	\caption{Test accuracy for various networks and their ensemble}
	\centering
	\resizebox{0.47\textwidth}{!}{%
		\begin{tabular}{lcccccc}
			\hline
			\textbf{Architectures} & \multicolumn{1}{c}{\textbf{\begin{tabular}[c]{@{}c@{}}Bengali\\ Digits\end{tabular}}} & \multicolumn{1}{c}{\textbf{\begin{tabular}[c]{@{}c@{}}Devanagari\\ Digits\end{tabular}}} & \multicolumn{1}{c}{\textbf{\begin{tabular}[c]{@{}c@{}}Telugu\\ Digits\end{tabular}}} & \multicolumn{1}{c}{\textbf{\begin{tabular}[c]{@{}c@{}}Bengali \\ Basic \\ Characters\end{tabular}}} & \multicolumn{1}{c}{\textbf{\begin{tabular}[c]{@{}c@{}}Bengali \\ Compound\\ Characters\end{tabular}}} &
			\multicolumn{1}{c}{\textbf{\begin{tabular}[c]{@{}c@{}}Mean \\ (Standard \\ Deviation)\end{tabular}}}	 \\ \hline \hline
			\textbf{LeNet} & 94.6 & 92.1 & 95.8 & 63.0 & 75.4 & 84.18($\pm$14.42) 
			\\
			\textbf{AlexNet} & 97.65 & 96.3 & 97.4 & 95.9 & 94.2& 96.29($\pm$1.39)
			\\
			\textbf{CapsNet} & 97.35 & 94.8 & 96.2 & 90.6 & 79.3& 91.65($\pm$7.36)
			\\
			\textbf{LeNet+AlexNet} & 97.6 & 96.2 & 97.3 & 95.8 & 92.6 &95.9($\pm$1.99)
			\\
			\textbf{LeNet+CapsNet} & 95.45 & 94 & 96.2 & 88.4 & 79.9 &90.79($\pm$6.81)
			\\
			\textbf{AlexNet+CapsNet} & \textbf{97.75} & \textbf{96.6} & 97.6 & \textbf{96.2} & \textbf{94.4}& \textbf{96.51($\pm$1.35)}
			\\
			\textbf{All\_Combined} & 97.5 & 96.5 & \textbf{97.8} & 96.1 & 75.2& 92.62($\pm$9.76)
			\\ \hline
		\end{tabular}%
	}
\end{table}
\begin{table}[h]
	\centering
	\caption{Comparitive study against other state of the art approaches}
	\resizebox{0.47\textwidth}{!}{%
		\begin{tabular}{llcll}
			\hline
			\textbf{Dataset} & \textbf{Our approach} & \textbf{Accuracy} & \textbf{Other approaches} & \textbf{Accuracy} \\ \hline \hline
			\multirow{3}{*}{\textbf{Bangla Digits}} & \multirow{3}{*}{\begin{tabular}[c]{@{}l@{}}AlexNet  + \\ CapsNet\end{tabular}} & \multirow{3}{*}{\textbf{97.75}} & Basu et al~\cite{basu2012mlp}. & 96.67 \\  
			&  &  & Roy et al.~\cite{roy2012new} & 95.08 \\ 
			&  &  & Roy et al.~\cite{roy2014axiomatic} & 97.45 \\ \hline
			\multirow{2}{*}{\textbf{\begin{tabular}[c]{@{}l@{}}Devanagari \\ Digits\end{tabular}}} & \multirow{2}{*}{\begin{tabular}[c]{@{}l@{}}AlexNet + \\ CapsNet\end{tabular}} & \multirow{2}{*}{\textbf{96.60}} & Das et al.~\cite{das2010handwritten} & 90.44 \\  
			&  &  & Roy et al.~\cite{roy2014axiomatic} & 96.50 \\ \hline
			\multirow{2}{*}{\textbf{\begin{tabular}[c]{@{}l@{}}Telugu \\ Digits\end{tabular}}} & \multirow{2}{*}{\begin{tabular}[c]{@{}l@{}}AlexNet  + \\ CapsNet + LeNet\end{tabular}} & \multirow{2}{*}{\textbf{97.80}} & Sarkhel et al.~\cite{sarkhel2015enhanced} & 97.50 \\ 
			&  &  & Roy et al.\cite{roy2014axiomatic} & 87.20\\ \hline
			\multirow{2}{*}{\textbf{\begin{tabular}[c]{@{}l@{}}Bangla Basic \\ Characters\end{tabular}}} & \multirow{2}{*}{\begin{tabular}[c]{@{}l@{}}AlexNet  + \\ CapsNet\end{tabular}} & \multirow{2}{*}{\textbf{96.20}} & Sarkhel et al.~\cite{sarkhel2015enhanced} & 86.53 \\ 
			&  &  & Bhattacharya et al.~\cite{bhattacharya2006recognition} & 92.15 \\ \hline
			\multirow{3}{*}{\textbf{\begin{tabular}[c]{@{}l@{}}Bangla \\ Compound\\ Characters\end{tabular}}} & \multirow{3}{*}{\begin{tabular}[c]{@{}l@{}}AlexNet + \\ CapsNet\end{tabular}} & \multirow{3}{*}{\textbf{94.40}} & Roy et al.~\cite{roy2017handwritten} & 90.33 \\ 
			&  &  & Pal et al.~\cite{pal2015recognition} & 93.12 \\
			&  &  & Sarkhel et al.~\cite{sarkhel2016multi} & 86.64 \\ \hline
		\end{tabular}%
	}
\end{table}
\subsection{Result and Analysis}
The result of the experiments have been tabulated in table 2. It can be clearly seen that capsule networks(written as CapsNet) surpasses LeNet in case of every dataset used. AlexNet being almost 7 times larger network as compared to capsule networks performs better than capsule networks. However the difference in performance is much more visible in case of the character datasets with much higher number of classes as compared to digits. LeNet fail poorly for the character datasets. Capsule network proved to be much more robust against complex data with higher number of classes. Upon combination we can see that combining LeNet with AlexNet is detrimental in nature with respect to AlexNet alone for every dataset. However combining capsule networks have always shown a positive effect. This proves that capsule network are capable of extracting some information that even AlexNet fails to obtain. For most datasets the best performance was achieved by combining AlexNet with capsule networks except for Telugu digits, where combination of all three networks proved to be the best. Furthermore we have analyzed the rise of test accuracy with every epoch of training. It can be seen that the capsule networks have the steepest slope signifying that they have the fastest learning curve. Finally in table 3 we have compared the obtained result against some state of the art works performed on the datasets. \par In terms of computational complexity, the extra computational overhead is during the dynamic routing phase. Other than training $W^{DC}$ by backpropagation, for every sample the routing weights $W^{DC}$ must also be tuned for $R_{iter}$ times, where $R_{iter}$ is the number of routing iterations. During each iteration $N_{PC}*N_{class}$ number of coupling coefficients must be calculated. Further an weighted sum over $N_{PC}$ dimension is needed to compute the combined digit capsule. Finally $N_{PC}*N_{class}$ number of routing weights must be tuned using agreement of individual and combined digit capsules. With all these, capsule networks generally have quite slow iterations, but as evident from Fig. \ref{fig:3} it also learns much faster as compared to LeNet and AlexNet.

\section{Conclusion}
In our current work we have implemented the capsule networks on handwritten Indic digits and character databases. We have shown that capsule networks are much superior and robust compared to the LeNet architecture. We have also seen that capsule networks can act as a booster when combined with other networks like LeNet and AlexNet. The best performance was achieved by combining AlexNet with capsule networks for most of the datasets. Only in case of Telugu dataset, combination of all three networks worked the best. From the results it can be concluded that even with 7 times more parameters that capsule networks, the AlexNet failed to capture some information that the capsule network learnt. Thus it was able to improve the performance of AlexNet. Finally it has also been seen the capsule network converge much faster that LeNet or AlexNet. In terms of pros and cons, the use of capsule networks can be beneficial for learning with much lesser number of features and also as improvement technique for other bigger networks. The problem with capsule network is its slow iterative process and limitation to single layer routing. That reveals many avenues of research. 

\section*{Acknowledgment}
This work is partially supported by the project order no. SB/S3/EECE/054/2016, dated 25/11/2016, sponsored by SERB (Government of India) and carried out at the Centre for Microprocessor Application for Training Education and Research, CSE Department, Jadavpur University.

\bibliographystyle{IEEETran}
\bibliography{ref}

\begin{thebibliography}{10}
\providecommand{\url}[1]{#1}
\csname url@samestyle\endcsname
\providecommand{\newblock}{\relax}
\providecommand{\bibinfo}[2]{#2}
\providecommand{\BIBentrySTDinterwordspacing}{\spaceskip=0pt\relax}
\providecommand{\BIBentryALTinterwordstretchfactor}{4}
\providecommand{\BIBentryALTinterwordspacing}{\spaceskip=\fontdimen2\font plus
\BIBentryALTinterwordstretchfactor\fontdimen3\font minus
  \fontdimen4\font\relax}
\providecommand{\BIBforeignlanguage}[2]{{%
\expandafter\ifx\csname l@#1\endcsname\relax
\typeout{** WARNING: IEEEtran.bst: No hyphenation pattern has been}%
\typeout{** loaded for the language `#1'. Using the pattern for}%
\typeout{** the default language instead.}%
\else
\language=\csname l@#1\endcsname
\fi
#2}}
\providecommand{\BIBdecl}{\relax}
\BIBdecl

\bibitem{lenet}
Y.~LeCun, L.~Bottou, Y.~Bengio, and P.~Haffner, ``Gradient-based learning
  applied to document recognition,'' \emph{Proceedings of the IEEE}, vol.~86,
  no.~11, pp. 2278--2324, 1998.

\bibitem{alexnet}
A.~Krizhevsky, I.~Sutskever, and G.~E. Hinton, ``Imagenet classification with
  deep convolutional neural networks,'' pp. 1097--1105, 2012.

\bibitem{routing}
S.~Sabour, N.~Frosst, and G.~E. Hinton, ``Dynamic routing between capsules,''
  pp. 3856--3866, 2017.

\bibitem{indic3}
S.~Ukil, S.~Ghosh, S.~M. Obaidullah, K.~Santosh, K.~Roy, and N.~Das, ``Deep
  learning for word-level handwritten indic script identification,''
  \emph{arXiv preprint arXiv:1801.01627}, 2018.

\bibitem{indic6}
R.~Sarkhel, N.~Das, A.~Das, M.~Kundu, and M.~Nasipuri, ``A multi-scale deep
  quad tree based feature extraction method for the recognition of isolated
  handwritten characters of popular indic scripts,'' \emph{Pattern
  Recognition}, vol.~71, pp. 78--93, 2017.

\bibitem{basu2012mlp}
S.~Basu, N.~Das, R.~Sarkar, M.~Kundu, M.~Nasipuri, and D.~K. Basu, ``An mlp
  based approach for recognition of handwritten bangla numerals,'' \emph{arXiv
  preprint arXiv:1203.0876}, 2012.

\bibitem{roy2012new}
A.~Roy, N.~Mazumder, N.~Das, R.~Sarkar, S.~Basu, and M.~Nasipuri, ``A new quad
  tree based feature set for recognition of handwritten bangla numerals,'' pp.
  1--6, 2012.

\bibitem{roy2014axiomatic}
A.~Roy, N.~Das, R.~Sarkar, S.~Basu, M.~Kundu, and M.~Nasipuri, ``An axiomatic
  fuzzy set theory based feature selection methodology for handwritten numeral
  recognition,'' pp. 133--140, 2014.

\bibitem{das2010handwritten}
N.~Das, B.~Das, R.~Sarkar, S.~Basu, M.~Kundu, and M.~Nasipuri, ``Handwritten
  bangla basic and compound character recognition using mlp and svm
  classifier,'' \emph{arXiv preprint arXiv:1002.4040}, 2010.

\bibitem{sarkhel2015enhanced}
R.~Sarkhel, A.~K. Saha, and N.~Das, ``An enhanced harmony search method for
  bangla handwritten character recognition using region sampling,'' pp.
  325--330, 2015.

\bibitem{bhattacharya2006recognition}
U.~Bhattacharya, M.~Shridhar, and S.~K. Parui, ``On recognition of handwritten
  bangla characters,'' pp. 817--828, 2006.

\bibitem{roy2017handwritten}
S.~Roy, N.~Das, M.~Kundu, and M.~Nasipuri, ``Handwritten isolated bangla
  compound character recognition: A new benchmark using a novel deep learning
  approach,'' \emph{Pattern Recognition Letters}, vol.~90, pp. 15--21, 2017.

\bibitem{pal2015recognition}
A.~Pal and J.~Pawar, ``Recognition of online handwritten bangla characters
  using hierarchical system with denoising autoencoders,'' pp. 0047--0051,
  2015.

\bibitem{sarkhel2016multi}
R.~Sarkhel, N.~Das, A.~K. Saha, and M.~Nasipuri, ``A multi-objective approach
  towards cost effective isolated handwritten bangla character and digit
  recognition,'' \emph{Pattern Recognition}, vol.~58, pp. 172--189, 2016.

\end{thebibliography}
\end{document}